\documentclass[10pt,twocolumn,letterpaper]{article}

\usepackage[pagenumbers]{cvpr} 

\usepackage{graphicx}
\usepackage{capt-of}
\usepackage{booktabs}
\usepackage{multirow}
\usepackage{colortbl}

\definecolor{cvprblue}{rgb}{0.21,0.49,0.74}
\usepackage[pagebackref,breaklinks,colorlinks,allcolors=cvprblue]{hyperref}

\def\confName{CVPR}
\def\confYear{2027}

\title{GTA: Generative Tower of Agents for Scalable Holonic Scene Generation}

\author{
Zeng Tao\textsuperscript{1},
Yunuo Chen\textsuperscript{1,2},
Xiang Feng\textsuperscript{1},
Yumeng He\textsuperscript{1},
Peihao Li\textsuperscript{2,3}\\[5pt]
Nanxiang Jiang\textsuperscript{1},
Yaowei Guo\textsuperscript{1},
Fanlian Zeng\textsuperscript{1,2},
Yin Yang\textsuperscript{4},
Chenfanfu Jiang\textsuperscript{1}\\[6pt]
{\small
\textsuperscript{1}UCLA\quad
\textsuperscript{2}Envora\quad
\textsuperscript{3}UC Berkeley\quad
\textsuperscript{4}University of Utah}
}

\begin{document}
\twocolumn[{
\renewcommand\twocolumn[1][]{#1}
\maketitle
\begin{minipage}{\textwidth}
  \centering
  \IfFileExists{fig/teaser.pdf}{%
    \includegraphics[width=\linewidth]{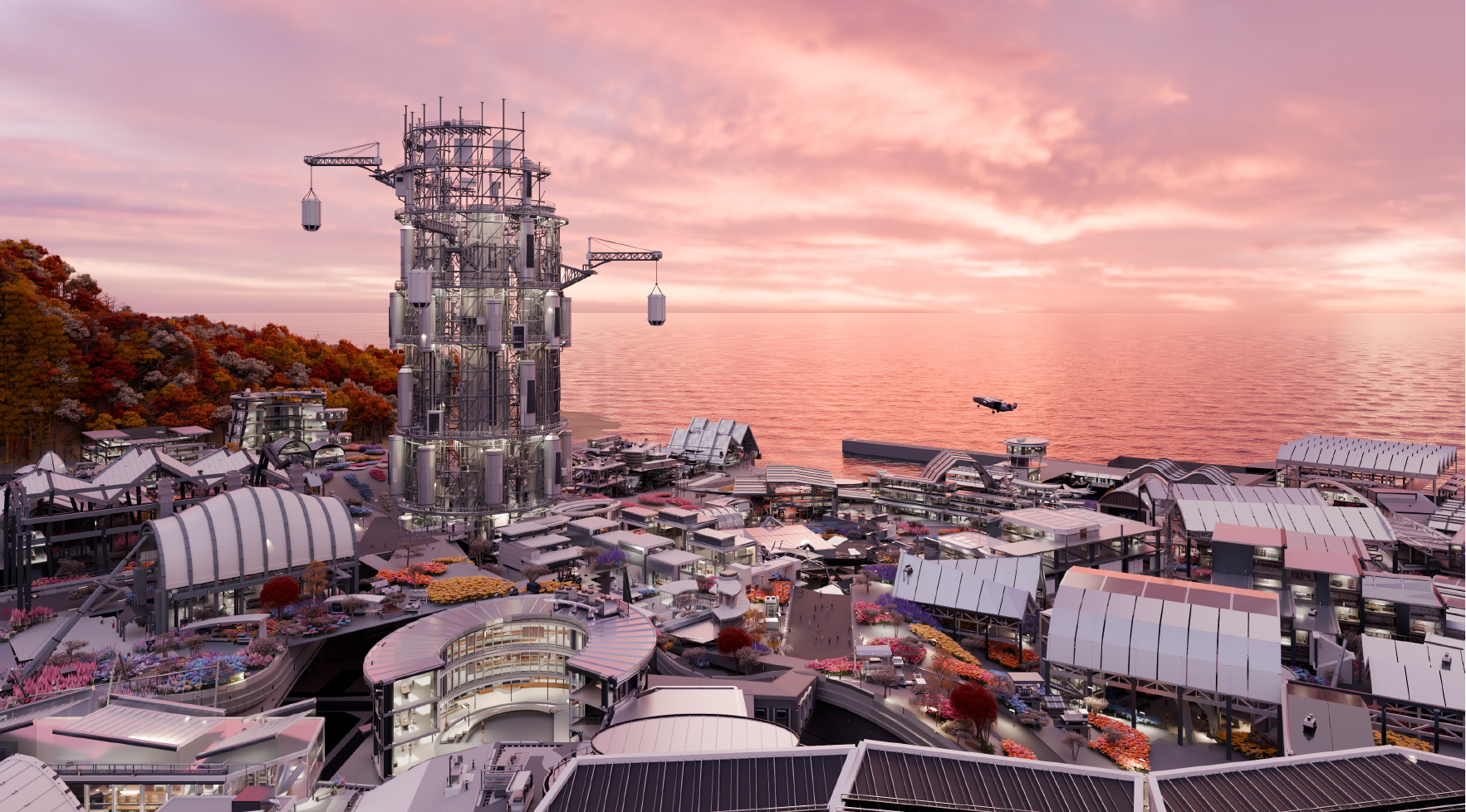}%
  }{%
    \fbox{\parbox[c][0.32\textheight][c]{\dimexpr\linewidth-2\fboxsep-2\fboxrule\relax}{%
      \centering\Large\textbf{GTA}\par\medskip
      \large Teaser: \textit{Tower of Babel} Construction Scene\par\medskip
      \normalsize Holon $\rightarrow$ Holon $\rightarrow$ Holon $\rightarrow$ Holon\par\smallskip
      Unified recursive construction across types and scales\par\medskip
      \small\textit{Teaser image placeholder (fig/teaser.pdf)}%
    }}%
  }
  \captionof{figure}{
  Building a \textit{Tower of Babel} in the AI era would require collaboration across humanity and a system to coordinate that vast effort efficiently. Likewise, \textbf{GTA} unifies scene representation across scales and types, which enables more agents to work in parallel as the scene scales up, making world-scale modeling feasible.}
  \label{fig:teaser}
\end{minipage}
\vspace{1em}
}]
\begin{abstract}
Scalable world modeling requires coordinating scene-wide goals with fine-grained objects and dynamic characters under limited time and context budgets. We introduce GTA (Generative Tower of Agents), which represents every unit at every level of an adaptive scene hierarchy as a holon, both a self-contained construction task and a composable scene unit. Parallel agents recursively plan the hierarchy top-down, generate assets, and assemble and refine them bottom-up. Uniform interfaces carry scene-wide requirements down the hierarchy, guiding local construction so that each holon fits into the larger one. GTA accelerates scalable world modeling through high-throughput parallel execution while preserving well-structured layouts and fine-grained asset details. GTA then populates the assembled environments with dynamic human characters whose appearance and motion are guided by the scene, yielding realized 4D worlds. \textbf{Project page:} \url{https://zt-zeng-tao.github.io/GTA/}
\end{abstract}
    
\section{Introduction}
\label{sec:intro}

\begin{flushright}
\begin{minipage}{0.95\columnwidth}
\small\raggedleft
\textit{``Parts and wholes in an absolute sense do not exist in the domains of life.''}\par
\medskip
--- Arthur Koestler, 1969~\cite{koestler1969holon}
\end{minipage}
\end{flushright}

Recent advances in agentic scene generation have enabled agents to construct complex 4D worlds through semantic reasoning, planning, and tool use~\cite{yang2026sceneweaver,xia2026sage,pfaff2026scenesmith,hollis2026sceneagent}. Rather than relying on rigid generation pipelines, agents can interpret open-ended instructions, orchestrate heterogeneous tools, and iteratively refine scene geometry and spatial relationships. This flexibility offers a promising path toward open-world scene generation.

However, agentic capabilities alone do not guarantee scalability across arbitrary scene types. As scenes grow more complex, scene generation must reconcile scene-wide semantic goals with spatial constraints while preserving fine-grained object detail. It must also place objects precisely to avoid interpenetration and ensure stable support for simulation within limited time and context budgets. Consider the construction of the \textit{Tower of Babel} in Fig.~\ref{fig:teaser}: agents must coordinate the layout of a production district, place workbenches within workshops, and model details as small as the teeth of a handsaw. Existing approaches often encode this process in predefined hierarchies or specialized sequential workflows, limiting their flexibility in open-world settings~\cite{deng2025global,li2026hdsl,guo2026worldclaw}, making it difficult to scale to larger scenes, preserve fine-grained object detail, and adapt to open-world settings. This raises a fundamental question: How can agents autonomously organize and construct scenes across diverse types and scales through a unified paradigm?

We introduce \textbf{GTA}, a \textbf{G}enerative \textbf{T}ower of \textbf{A}gents for scalable scene construction built upon a unified parallel recursive framework. GTA is guided by three core agentic insights. (1) Adaptivity, enabling agents to determine scene organization and decomposition based on open-ended inputs. (2) Uniformity, allowing agents to reuse the same reasoning and tool-use procedures across scene units. (3) Interoperability, enabling agents to pass constraints down and construction results up through uniform interfaces.

At the core of GTA is the \textbf{Holon}. Arthur Koestler introduced the concept of the holon in \textit{The Ghost in the Machine}~\cite{koestler1967ghost}. In GTA, a holon is a construction unit that is both a whole in itself and a part of a larger scene. Each Holon forms a self-contained construction task while remaining a composable component of a larger Holon. A world is a Holon, just as a city is a Holon, a park is a Holon, a flowerbed is a Holon, and a plant is a Holon. What changes across scales is the content of the Holon, not the paradigm by which it is constructed. This unified abstraction, named \textbf{Holarchy}, allows agents to recursively organize diverse scenes without predefined hierarchies.

Building on Holarchy, GTA establishes a unified recursive agentic scene generation paradigm consisting of Top-down Planning, Asset Generation, and Bottom-up Placement. \textbf{Top-down Planning} decomposes the scene specification into a hierarchy of holons, \textbf{Asset Generation} produces the corresponding assets, and \textbf{Bottom-up Placement} assembles these assets within their holon groups and recursively integrates the groups into a complete scene. The recursive structure of the holarchy and its uniform interfaces allow independent branches to be processed in parallel at each stage, preserving fine-grained asset detail while respecting scene-level semantic constraints. A post-generation stage then enriches the resulting 4D world with high-quality dynamic human characters whose appearance and motion suit the scene's spatiotemporal semantics.

We evaluate GTA across diverse scene types, spatial scales, and levels of semantic complexity, examining scene quality, generation efficiency, and the scalability of recursive and parallel construction.

Our main contributions are summarized as follows:

\begin{itemize}
\item We introduce a unified scene unit representation that enables agents to adaptively organize heterogeneous scenes without scene-specific templates, supporting a consistent agentic workflow across types and scales.
\item We develop a recursive framework integrating top-down planning, asset generation, and bottom-up placement. By combining reusable skills with parallel execution of independent Holons, our approach enables scalable and accelerated scene construction.
\item We evaluate GTA across a range of scene types and scales, showing that high-throughput parallel execution speeds up scalable world modeling while maintaining coherent layouts and detailed assets, as well as 4D worlds populated with scene-guided animated characters.
\end{itemize}

\begin{figure*}[th]
    \centering
    \includegraphics[width=\linewidth]{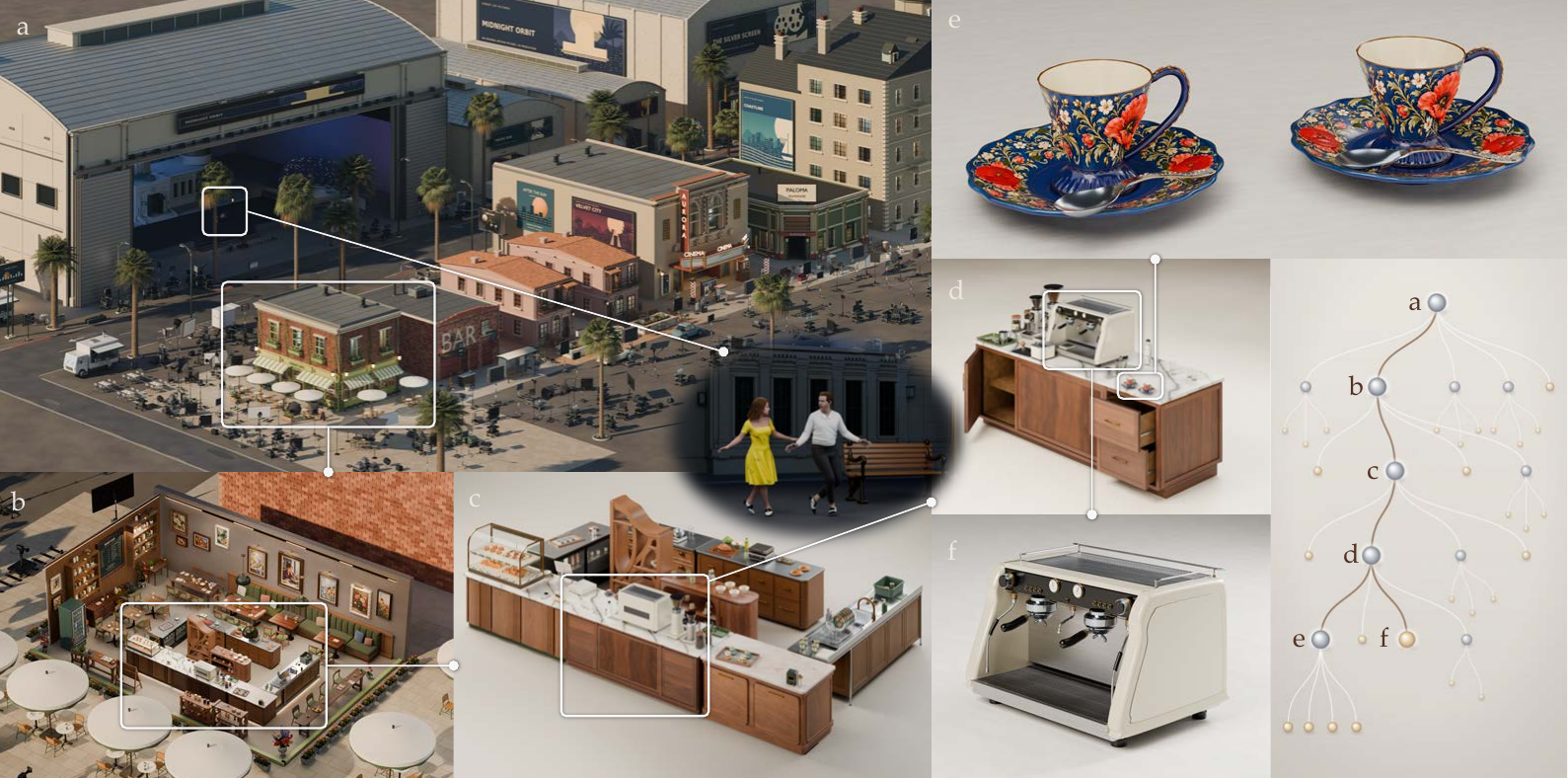}
    \caption{\textbf{A film studio backlot with holarchy.} GTA recursively decomposes the backlot (a) into holons such as a cafe (b), which contains a counter area (c) and smaller components such as a coffee station (d), cups and saucers (e), and an espresso machine (f). The labels connect each rendered example to its corresponding holon on the right.}
    \label{fig:method}
    \vspace{-0.5cm}
\end{figure*}

\section{Related Work}
\label{sec:related_work}




\subsection{3D Scene Generation}

Early learning-based methods~\cite{yu2011make,tang2024diffuscene} such as ATISS~\cite{paschalidou2021atiss} synthesize indoor layouts autoregressively from room types and floor plans. Recent approaches~\cite{gao2023scenehgn,lin2024instructscene,gao2024graphdreamer,chen2023scenedreamer} offer greater control over scene structure. HOG-Layout~\cite{jiang2026hog} uses a hierarchical representation and vision-language reasoning to generate, optimize, and edit indoor layouts. WorldGen~\cite{wang2026worldgen} combines procedural layout planning with holistic image-to-3D reconstruction, then decomposes the result into individual meshes for object-level refinement. GTA recursively decomposes each scene into a hierarchy of holons, adapting its depth and branching to the scene structure. The hierarchy specifies which components to construct and how they are nested within the scene.

\subsection{Agentic Scene Generation}

Language-guided systems use foundation models to infer scene content and coordinate specialized tools~\cite{hu2024scenecraft,deng2025global,yang2026sceneweaver,guo2026worldclaw,li2026hdsl,liu2026scene}. Holodeck~\cite{yang2024holodeck} retrieves assets from Objaverse and optimizes their placement under spatial constraints proposed by an LLM. SAGE~\cite{xia2026sage} generates simulation environments for embodied tasks and iteratively refines them using visual and physical feedback. SceneSmith~\cite{pfaff2026scenesmith} uses designer, critic, and orchestrator agents across specialized indoor construction stages and parallelizes work across rooms. From visual captures, SceneAgent~\cite{hollis2026sceneagent} uses simulator-based inspection and repair to turn reconstructed scenes into interactive environments. GTA treats holon groups at every level as uniform units of agentic execution. Through common interfaces, agents reuse the same operations across levels and process groups in parallel as their dependencies are resolved, accelerating scene construction.

\subsection{Compositional and Multi-Agent Systems}
Agent systems coordinate subtasks~\cite{wu2023autogen,hong2024metagpt} and schedule execution by dependency~\cite{kim2023llm,yu2025dyntaskmas}. ADaPT~\cite{prasad2024adapt} decomposes subtasks on demand, while ROMA~\cite{alzu2026roma} combines recursive decomposition, parallel execution, and bottom-up aggregation. GTA ties each construction task to a spatial component, passing constraints down the hierarchy and placing completed child holons within their parents.

\section{Method}
\label{sec:method}

\subsection{Overview}

Given scene text prompts, GTA outputs a detailed 3D scene constructed through a unified holon representation and parallel recursive execution, together with scene-guided 4D humans. We represent the scene as holons as shown in Fig.~\ref{fig:method} and organize agentic operations around holon groups, with shared interfaces propagating constraints across levels (Sec.~\ref{sec:holon}). Construction proceeds in three stages: top-down planning recursively decomposes the scene; asset generation in parallel creates assets for leaf holons; and bottom-up placement and refinement assembles highly detailed components (Sec.~\ref{sec:parallel}). Finally, a scene-guided human generation pipeline creates and integrates dynamic characters with appropriate activities and appearance (Sec.~\ref{sec:human}).

\subsection{Holon Representation for Agents}
\label{sec:holon}

\begin{figure}[h]
    \centering
    \includegraphics[width=1\linewidth]{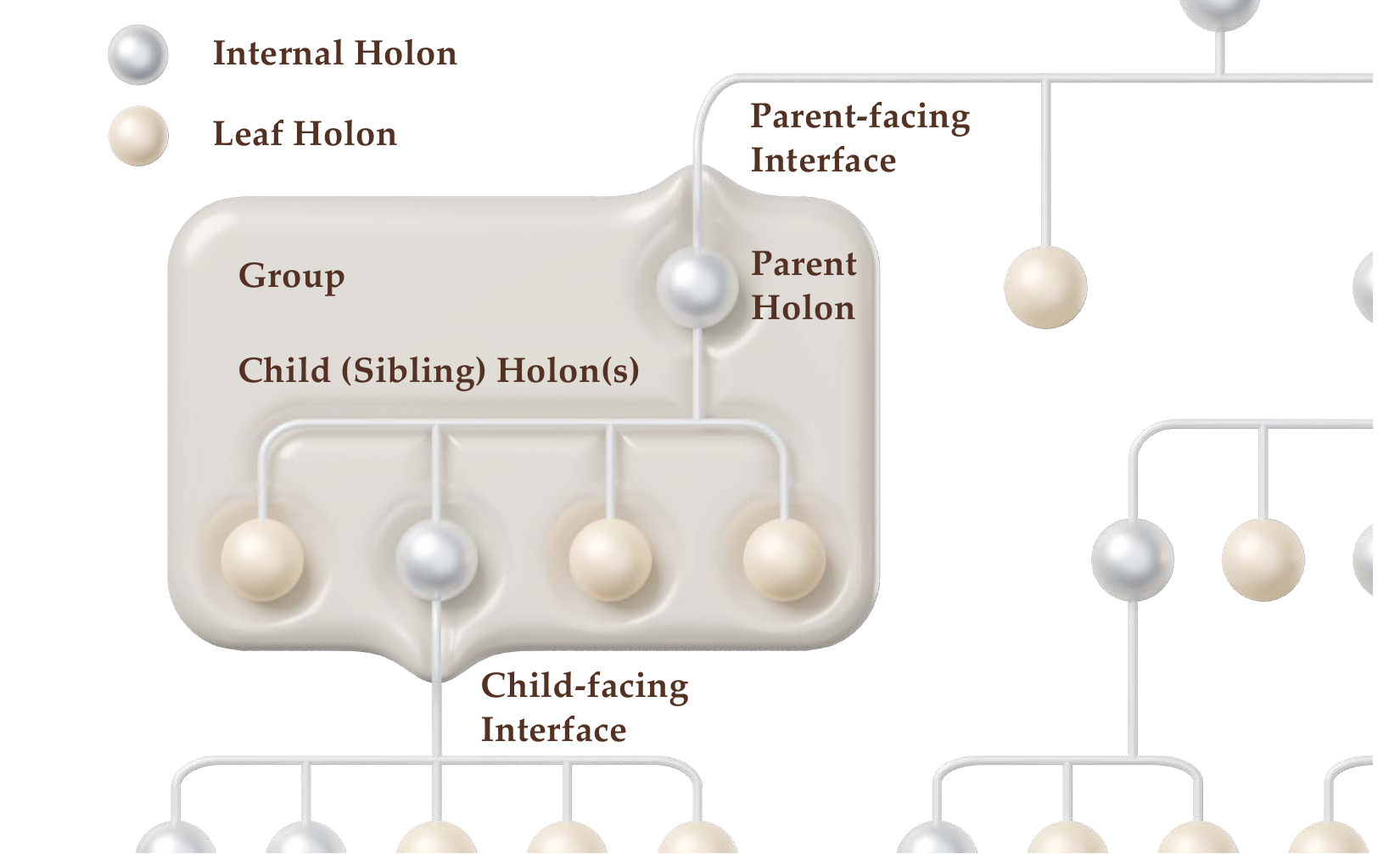}
    \caption{\textbf{Holons and their interfaces.} Each holon group consists of a parent holon and its immediate children and serves as the unit on which agents operate. Groups communicate with one another through parent-facing and child-facing interfaces.}
    \label{fig:pipeline_1}
    \vspace{-0.3cm}
\end{figure}

In GTA, a \textbf{holon} is the fundamental unit of scene processing.
A \textit{leaf holon} represents an indivisible object, while an
\textit{internal holon} represents a composition of child holons.
As shown in Fig.~\ref{fig:pipeline_1}, the entire scene forms the
\textit{root holon}. These parent--child relationships define the
holarchy as a rooted tree $\mathcal{T}=(\mathcal{H},E)$, where
$\mathcal{H}$ is the set of holons and $(h,c)\in E$ means that $c$
is an immediate child of $h$. The children of $h$ are siblings and
can themselves be leaf or internal holons. Agents operate on
\textit{holon groups}. For each internal holon $h$, its group contains
$h$ and its immediate children $\operatorname{Ch}(h)$:
\begin{equation}
    \mathcal{G}(h)=\{h\}\cup\operatorname{Ch}(h).
\end{equation}
Groups communicate through \textbf{parent-facing} and
\textbf{child-facing interfaces}. This recursive structure supports
a common workflow across scene types and scales. Groups without
interdependencies run in parallel across agentic slots, accelerating
generation.



\noindent\textbf{Unification.} At every level of the scene hierarchy, agents operate on holon groups using the same reasoning and tool-use procedures. Uniform interfaces pass constraints downward and construction results upward. Given the global context, each agent only needs to reason about relationships within its group and the interfaces that connect it to other groups. This local scope allows the same workflow to scale to larger scenes.

\begin{figure}[h]
    \centering
    \includegraphics[width=1\linewidth]{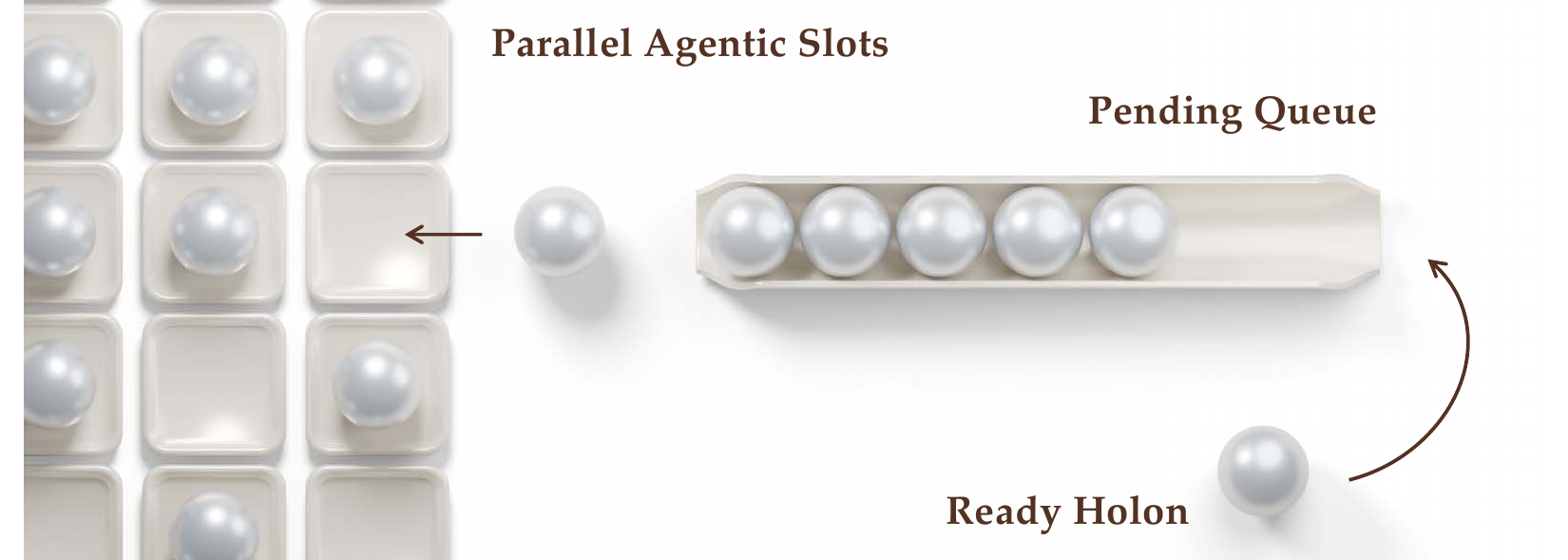}
    \caption{\textbf{Parallel agentic processing of holons.} Once a holon's dependencies are satisfied, it enters the pending queue. The scheduler dispatches queued groups to available agentic slots, allowing independent holons to execute concurrently. }
    \label{fig:pipeline_2}
\end{figure}

\noindent\textbf{Parallelism.} At each stage, the system maintains a queue of holons awaiting execution and a set of agentic slots, as shown in Fig.~\ref{fig:pipeline_2}. Once a holon group is ready for processing, it enters the queue. The scheduler continuously dispatches queued groups to available slots. GTA reduces the bottleneck of sequential construction and allows throughput to scale with system capacity, thus accelerating generation.

\vspace{-0.3cm}

\subsection{Parallel Recursive Agentic Execution}
\label{sec:parallel}

Building on the holon representation, our pipeline consists of three stages, each supporting parallel agent execution over the recursive scene structure. As the scene grows, the potential for concurrent processing also increases.

\noindent\textbf{Top-down Planning.} Stage 1 recursively plans the scene from the root holon, processing independent groups in parallel. For each internal holon, an agent only deals with the global and group-level local context to determine its child holons. Once the group has been planned, all internal child holons are added to the processing queue. Available agentic slots then process these holons concurrently. Planning is complete when the queue is empty and no planning tasks remain active.

\noindent\textbf{Asset Generation.} Stage 2 generates assets for all leaf holons. Each leaf holon represents an indivisible object with specified attributes, allowing its asset to be generated independently. Image guidance enriches visual detail, and text guidance is well suited to modeling assets according to explicit specifications. In Stage 2, GTA uses GPT-6 Astra~\cite{openai_gpt6_astra} and Blender for text-guided modeling, matching the specifications and dimensions set in Stage 1 to facilitate subsequent placement. In Stage 3, GTA uses image models~\cite{openai_gpt_image_15} to guide appearance refinement during asset placement.


\noindent\textbf{Bottom-up Placement and Refinement.} Stage 3 assembles the scene from the bottom up. The system maintains a queue of groups that are ready for placement. Each group has one of three states: \textit{not-ready}, \textit{ready}, or \textit{finished}, while leaf holons are treated as finished. A group becomes ready and enters the queue once all its children have finished. Consequently, placement begins with groups whose children are all leaf holons and progressively moves upward through the hierarchy. Agents process ready groups concurrently in the available slots. Alongside placement, refinement agents improve the visual quality of the assets. They first use image editing to enhance renderings of the initial assets produced in the second stage, then use the edited images as visual guidance to refine the assets’ details.

\begin{table*}[t]
\centering
\caption{\textbf{Quantitative comparison between different methods.} T/U denote relative absolute errors in total instance and distinct semantic type counts; $P_{1}$ and MPD measure penetration frequency and depth, respectively. Two visual judges assess object- and scene-level quality on a 1–5 scale. ↑/↓ indicates higher/lower is better. Within each group, the best and second-best results are highlighted in \textbf{bold} and \underline{underlined}, respectively. $\dagger$ Astra-O, Astra-S, and GTA use the same GPT-6 Astra backbone with medium reasoning effort.}
\label{tab:quantitative_results}
\small
\setlength{\tabcolsep}{4pt}
\resizebox{0.9\linewidth}{!}{%
\begin{tabular}{cccccccccc}
\toprule
& & \multicolumn{2}{c}{Prompt Compliance} & \multicolumn{4}{c}{Physical Validity} & \multicolumn{2}{c}{Visual Quality (Obj. / Scene)} \\
\cmidrule(lr){3-4}\cmidrule(lr){5-8}\cmidrule(lr){9-10}
\multirow{-2}{*}[7pt]{Scene Type} & \multirow{-2}{*}[7pt]{Method} & \raisebox{4.5pt}{Quantity Error (T / U)$\downarrow$ (\%)} & \shortstack{Specified\\Objects $\uparrow$(\%)} & \raisebox{4.5pt}{$P_1$ $\downarrow$ (\%)} & \shortstack{MPD $\downarrow$\\(mm)} & \raisebox{4.5pt}{NAV~\cite{tam2026sceneeval} $\uparrow$} & \shortstack{Opening\\Clearance~\cite{tam2026sceneeval} $\uparrow$} & \raisebox{4.5pt}{Claude Opus 5.5 $\uparrow$} & \raisebox{4.5pt}{GPT-6.1 Sol $\uparrow$} \\
\midrule
\multirow{3}{*}{\shortstack{Indoor\\Scenes}} & SAGE             & 32.92 / 31.38 & \underline{64.29–100.00} & 15.25 & 8.02 & 0.785 & 0.976 & 2.55 / 2.08 & 2.83 / 2.38 \\
& SceneSmith       & \underline{27.78} / \underline{17.50} & 55.03–100.00 & \underline{13.64} & \textbf{2.97} & \underline{0.979} & \textbf{0.988} & \underline{3.65} / \underline{2.50} & \textbf{3.86} / \underline{3.31} \\
\cmidrule(l){2-10}
\rowcolor{black!8}\cellcolor{white} & GTA (Ours)             & \textbf{11.67} / \textbf{15.82} & \textbf{96.30–100.00} & \textbf{6.45} & \underline{4.53} & \textbf{0.983} & \underline{0.987} & \textbf{3.74} / \textbf{3.17} & \underline{3.81} / \textbf{3.86} \\
\midrule
\multirow{5}{*}{\shortstack{General\\Scenes}} & Fable 5.2       & 13.58 / 22.82 & \underline{89.92–100.00} & \underline{15.29} & \textbf{4.58} & \underline{0.994} & \underline{0.992} & \underline{2.70} / 2.85 & \underline{3.02} / \underline{3.45} \\
& Astra-O$^{\dagger}$ & 7.88 / 27.97 & 84.06–100.00 & 24.38 & 16.70 & 0.983 & 0.989 & 2.37 / \textbf{2.94} & 2.87 / 3.42 \\
& Astra-S$^{\dagger}$   & \underline{5.02} / \underline{11.39} & 58.18–100.00 & 27.47 & 25.26 & 0.987 & 0.985 & 2.29 / 1.84 & 2.75 / 2.21 \\
\cmidrule(l){2-10}
\rowcolor{black!8}\cellcolor{white} & GTA (Ours)$^{\dagger}$             & \textbf{2.93} / \textbf{9.59} & \textbf{93.21–100.00} & \textbf{8.54} & \underline{6.28} & \textbf{0.998} & \textbf{0.994} & \textbf{3.49} / \underline{2.92}  & \textbf{3.78} / \textbf{3.63} \\
\bottomrule
\end{tabular}%
}
\vspace{-0.2cm}
\end{table*}

\subsection{Human Generation}
\label{sec:human}


After generating the 3D scene, we populate it with human characters through a scene-guided post-generation pipeline. We first determine their number, roles, activities, and appearance based on the scene’s function, regional context, and available space, and then generate reference images to guide character creation. The pipeline comprises two generation branches: the static branch employs NLF~\cite{sarandi2024neural} and SMPL-X~\cite{pavlakos2019expressive} to recover body shapes and poses from reference images, while the dynamic branch uses Kimodo~\cite{rempe2026kimodo} to generate motion sequences conditioned on textual descriptions and trajectories. DressAgent~\cite{jiang2026inverse} generates garments and textures, DiffLocks~\cite{rosu2025difflocks} produces hair, and Meshy 7.1~\cite{meshy71} and TRELLIS.2~\cite{xiang2026native} generate footwear. Reference-based facial texture mapping and skin-tone harmonization further enrich character-specific appearance details. These components are assembled around a common body representation, with garment simulation performed once the body geometry and pose are established. Finally, the characters are integrated into the scene, with adjustments to ensure proper ground contact, object interactions, and collision avoidance, as illustrated in Fig.~\ref{fig:method}.

\begin{figure}[h]
    \centering
    \includegraphics[width=\linewidth]{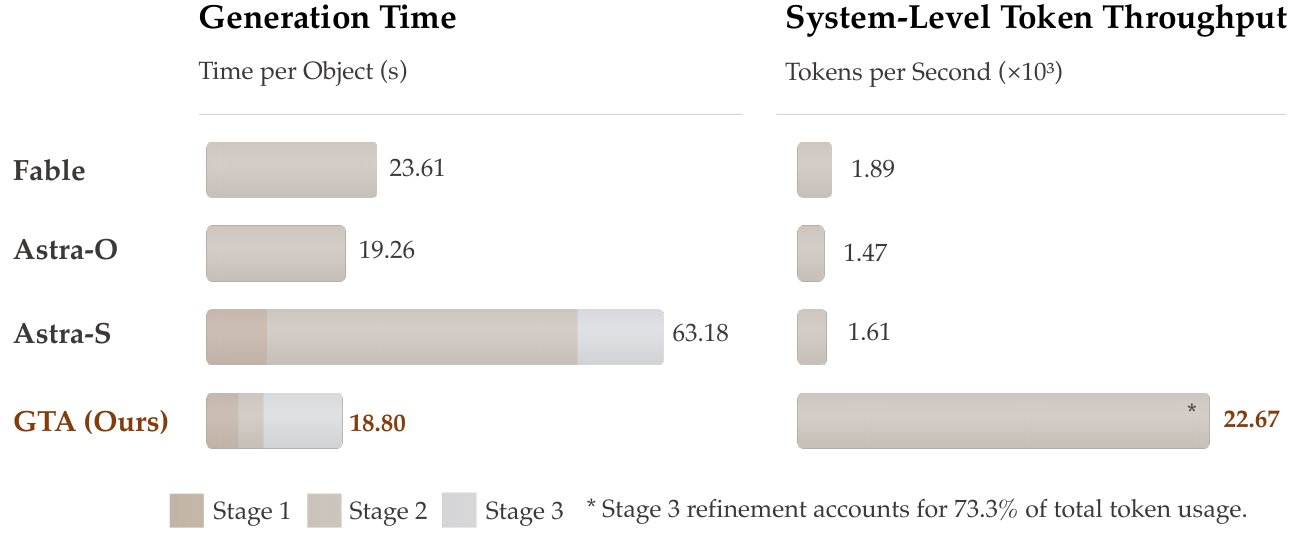}
    \caption{\textbf{Faster generation with higher agentic throughput.} The left panel shows wall-clock time per object. The right panel shows system-level token throughput.}
    \label{fig:cost}
    \vspace{-0.3cm}
\end{figure}

\begin{figure*}[h]
    \centering
    \includegraphics[width=0.94\linewidth]{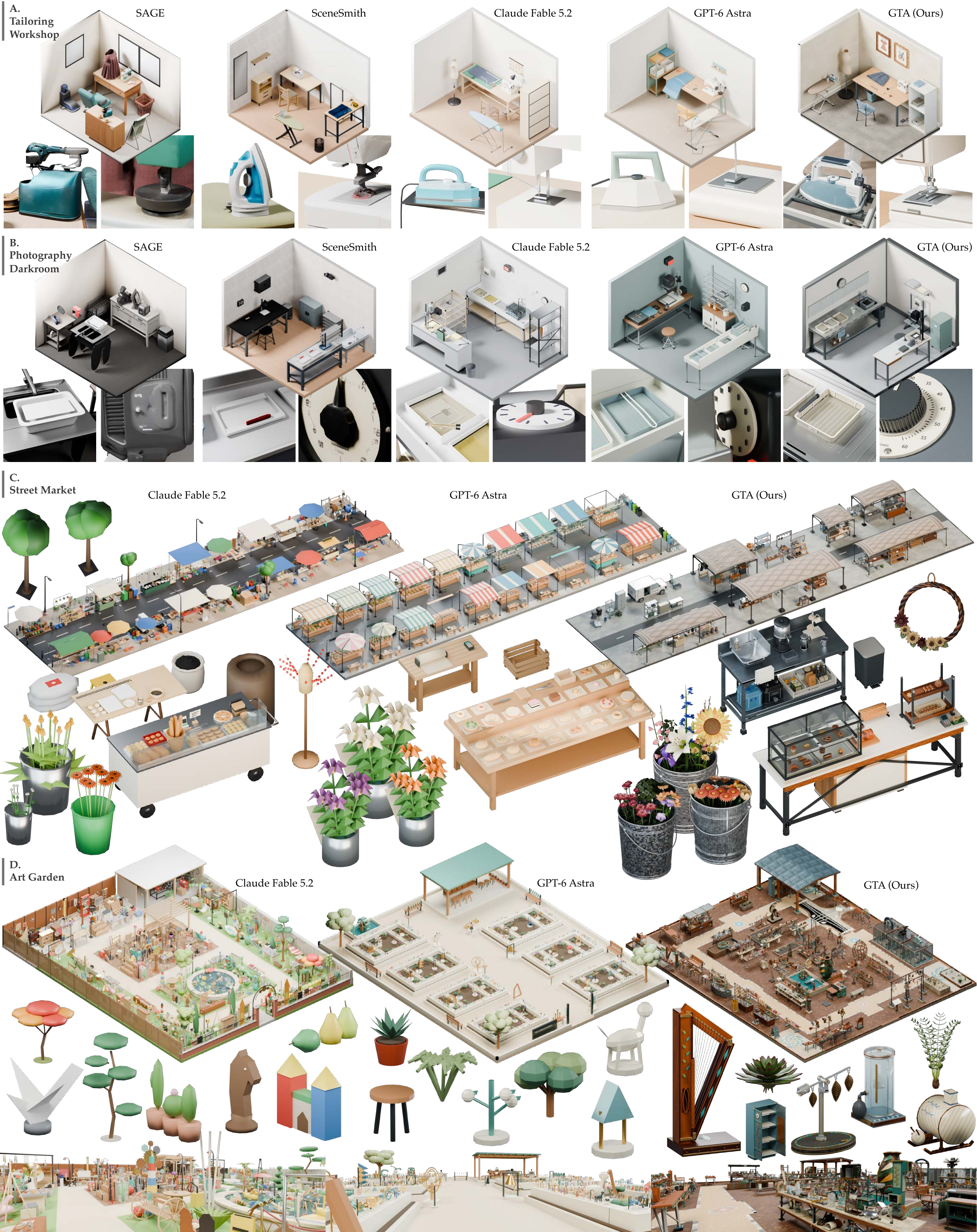}
    \caption{\textbf{Qualitative comparison across different scene types and scales.} We compare GTA with baselines on two indoor scenes (A--B) and two outdoor scenes (C--D). Full-scene views and object close-ups show how GTA combines coherent layouts with detailed, scene-specific assets, from darkroom equipment to market stalls and garden sculptures.}
    \label{fig:qual}
\end{figure*}

\begin{figure*}[h]
    \centering
    \includegraphics[width=0.95\linewidth]{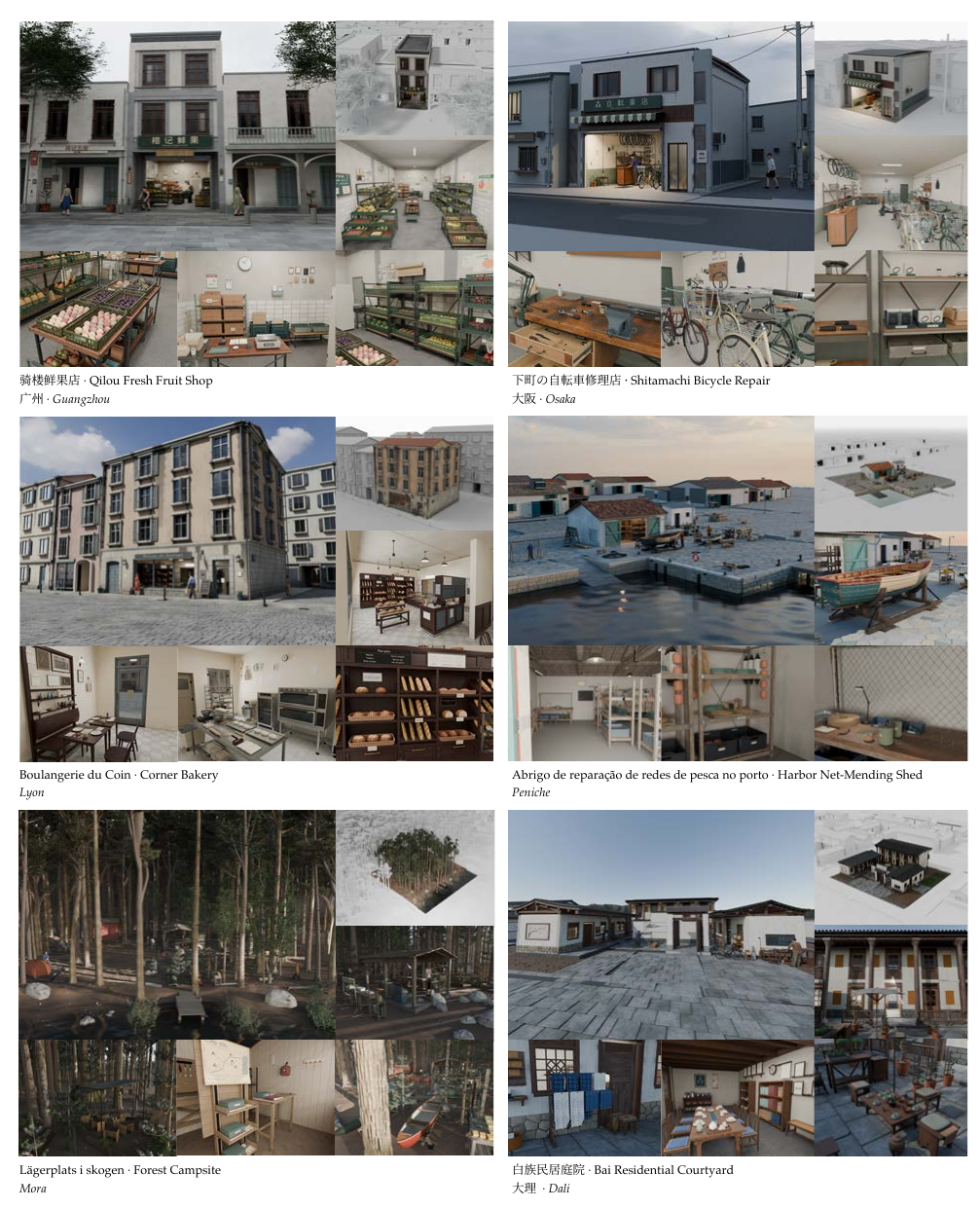}
    \caption{\textbf{Roaming the World with GTA.} GTA generates scenes across diverse geographic and cultural settings. Each scene consists of over 1,000 objects. GPT-6 Astra fills the gray-marked backgrounds, with some textures coming from external sources~\cite{polyhaven,polyhaven_sky_fire,cgbookcase_birch,mapzen_terrain}.  Please visit the \href{https://zt-zeng-tao.github.io/GTA/}{\textbf{Project Page}} for more high-resolution results.}
    \label{fig:addresult_1}
\end{figure*}

\begin{figure*}[h]
    \centering
    \includegraphics[width=0.97\linewidth]{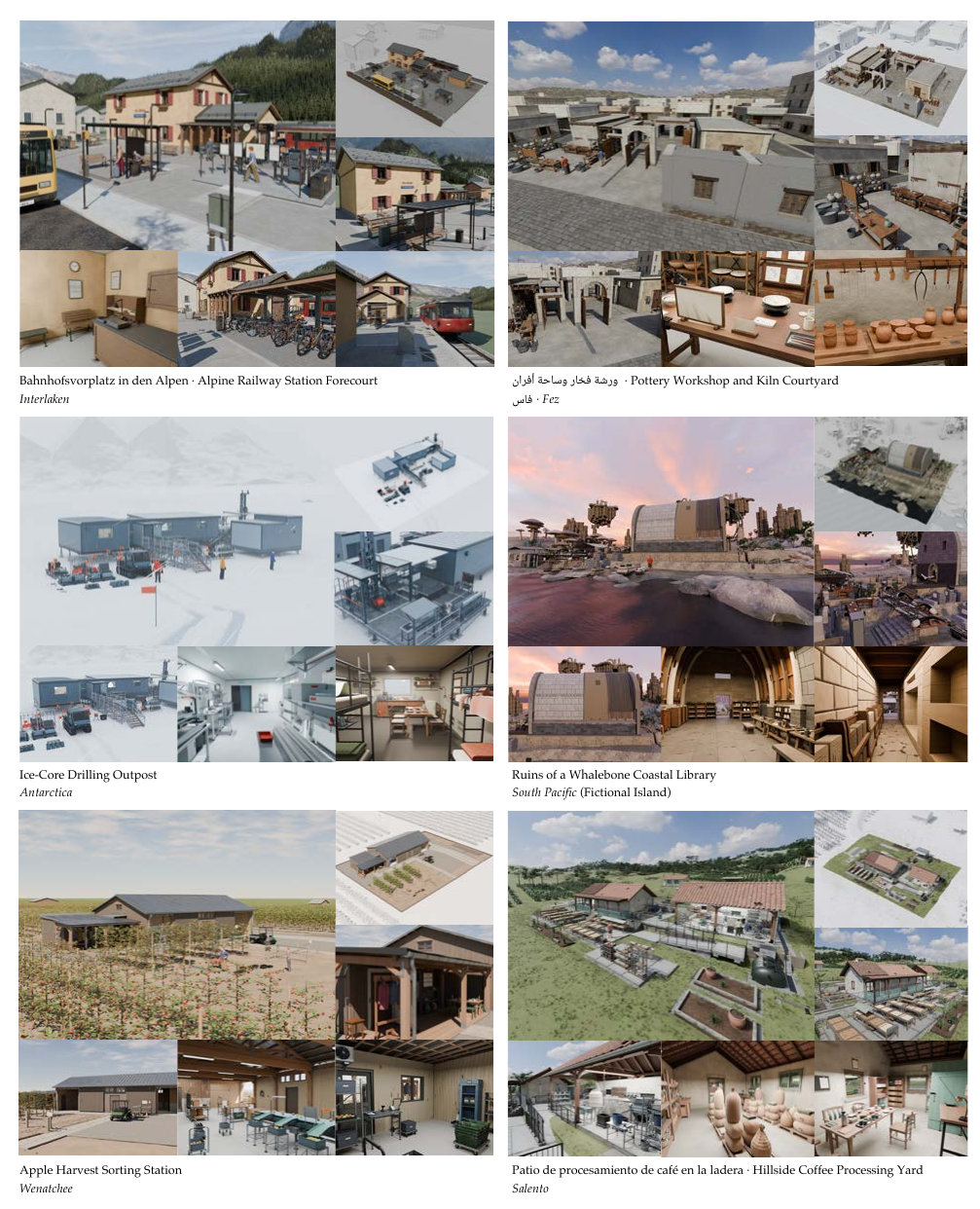}
    \caption{\textbf{Roaming the World with GTA (continued).} Additional scenes from the same series. Please visit the \href{https://zt-zeng-tao.github.io/GTA/}{\textbf{Project Page}} for more high-resolution results.}
    \label{fig:addresult_2}
\end{figure*}

\section{Experiments}
\label{sec:experiments}

\subsection{Experimental Setup}

GTA uses GPT-6 Astra~\cite{openai_gpt6_astra} with medium reasoning effort both to process holons through agents running in the Codex CLI and to make direct API calls. We use GPT Image 1.5~\cite{openai_gpt_image_15} for image generation. For the main experiments, we configure GTA with 32 parallel agentic slots. Each baseline runs a single model agent, and each GTA slot runs one model agent. We evaluate on a set of 20 indoor scenes and 20 additional outdoor or semi-outdoor scenes spanning a range of scales and types. Target counts range from 10 to 1,000 objects. All rendering throughout scene generation and evaluation uses Blender 4.5.0 with the Cycles engine and OptiX backend on the same set of NVIDIA H100 GPUs, with fixed parameters across methods within each evaluation group. We use Claude Opus 5.5~\cite{anthropic_claude_opus_55} and GPT-6.1 Sol~\cite{openai_gpt61_sol} for visual evaluation.

\subsection{Comparison with other methods}

We compare GTA with SAGE~\cite{xia2026sage} and SceneSmith~\cite{pfaff2026scenesmith} on indoor scenes, using GPT-6 Astra (medium reasoning effort) for language model calls and GPT Image 1.5 for image generation. On a broader set of scenes, we compare GTA with Claude Fable 5.2~\cite{anthropic_claude_fable_52} and GPT-6 Astra, both at medium reasoning effort. We evaluate the Original (Astra-O) and Staged (Astra-S) baselines, with Astra-S following the same three-stage workflow as GTA.

\noindent\textbf{Qualitative Comparison.} Fig.~\ref{fig:qual} compares GTA with several baselines across different scene types. The scene overviews and object close-ups highlight differences in layout, object detail, and physical plausibility. \textbf{(1) Richer, function-aware layouts.} In Fig.~\ref{fig:qual} C, Fable and Astra-O reuse similar spatial units, stall structures, and display arrangements. GTA lays out and furnishes each stall according to its function. The bakery stall displays different baked goods in a glass case, and the flower stall groups flowers of different shapes and colors in textured buckets. These variations make individual stalls easier to distinguish within a coherent street layout. \textbf{(2) Finer per-object detail.} Fig.~\ref{fig:qual} A and B show recognizable functional details, including the sewing machine’s needle and presser foot, the grooves in the developing tray, and the markings on an equipment dial. In D, Fable and Astra-O create scene-level variety largely through color and simplified geometry. GTA maintains clear paths and distinct exhibition areas while modeling details such as the strings and frame of the harp-shaped installation and the individual components of the hanging sculpture. \textbf{(3) More plausible physical relationships.} The close-ups reveal visible interpenetration in some baseline assets, including a vase in Fable’s result (Fig.~\ref{fig:qual} C). GTA’s corresponding examples show clearer object boundaries and more plausible contacts, consistent with the lower penetration rates reported in Table~\ref{tab:quantitative_results}.

\noindent\textbf{Quantitative Comparison.} We evaluate quantity compliance, physical plausibility, and visual quality. As shown in Table~\ref{tab:quantitative_results}, GTA achieves the lowest errors in both total instance count and distinct type count across indoor and general scenes. GTA also reduces the fraction of objects with penetration depths exceeding 1 mm ($P_1$), corresponding to relative reductions of 52.7\% and 44.1\% over the best baseline in each setting. GTA ranks second in mean penetration depth (MPD) in both groups, achieves the highest NAV score in both groups, and the highest opening-clearance score on general scenes. GTA receives the highest visual-quality score in six of the eight comparisons, suggesting that GTA can preserve geometric and material detail in individual objects while constructing complete scenes. With the GPT-6 Astra backbone held fixed, GTA outperforms Astra-O and Astra-S on different metrics.

\noindent\textbf{Speed and Throughput.} Fig.~\ref{fig:cost} shows that GTA achieves the lowest wall-clock time per object across all the compared methods. GTA achieves a 3.36$\times$ speedup over Astra-S in wall-clock time per object. Compared with Astra-O, GTA reduces time per object from 19.26 to 18.80 seconds and achieves 15.4$\times$ the system-level token throughput. Stage 3 refinement accounts for 73.3\% of GTA’s token usage, indicating that most of this capacity is devoted to refining individual objects. These results show that GTA’s parallel execution accommodates more per-object refinement within a shorter generation time.

\subsection{Ablation Studies}

\begin{figure}[h]
    \centering
    \includegraphics[width=\linewidth]{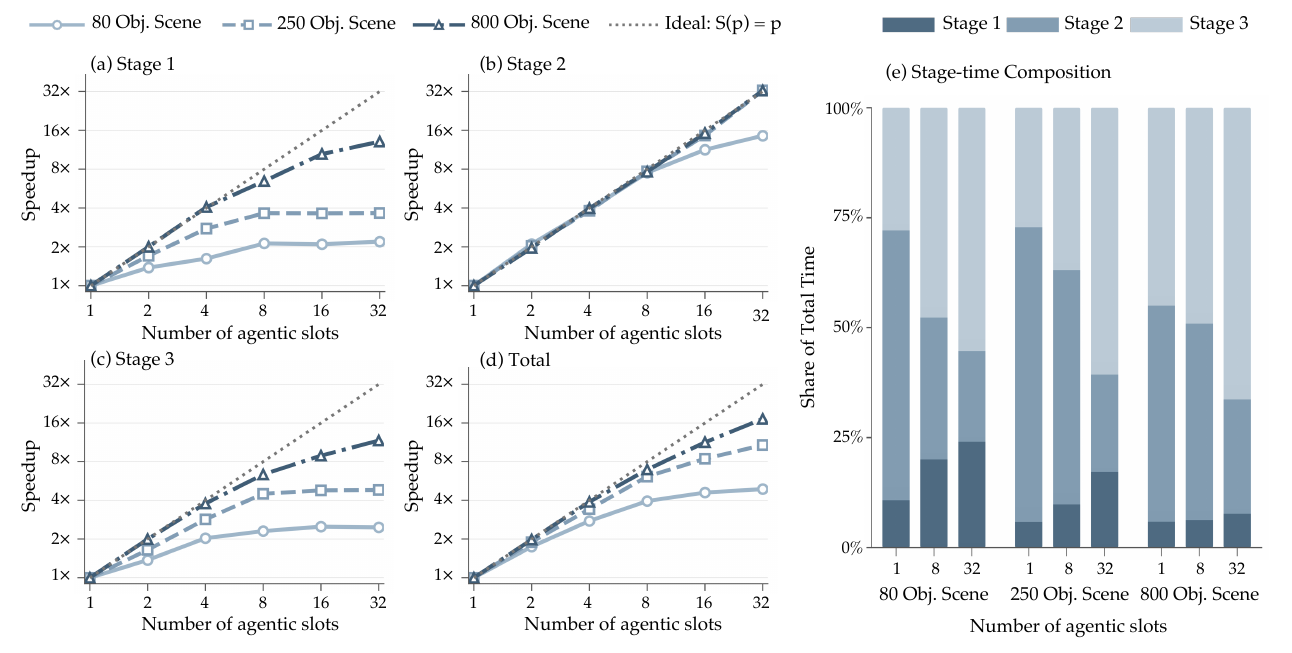}
    \caption{\textbf{Scaling up the agentic slot numbers speeds up scene generation.} We compare per-stage and end-to-end speedups for 1--32 slots with the ideal linear $S(p)=p$. The stacked bars show each stage's share of total runtime.}
    \vspace{-0.3cm}
    \label{fig:ablation}
\end{figure}

We evaluate the effect of the number of agentic slots on generation speed using three scenes containing 80, 250, and 800 objects. For each scene, we hold the input and experimental conditions fixed and measure the runtime of each stage with 1, 2, 4, 8, 16, and 32 slots. We run each configuration three times and report the mean runtime. We compute speedups relative to the single-slot runtime. As shown in Fig.~\ref{fig:ablation}, increasing the number of slots accelerates generation for all scenes. Larger scenes achieve higher end-to-end speedups. The 80-object scene reaches a plateau early, the 250-object scene attains a higher speedup, and the 800-object scene achieves the largest speedup at 32 slots. These results indicate that larger scenes can make better use of additional parallel resources. Stage 2 scales closest to the ideal linear speedup. As scene scale and slot count increase, Stage 3 becomes the dominant contributor to end-to-end runtime. At 32 slots, it accounts for more than half of the total runtime in each scene, making it the primary bottleneck for further speedup.

\begin{figure}[h]
    \centering
    \includegraphics[width=\linewidth]{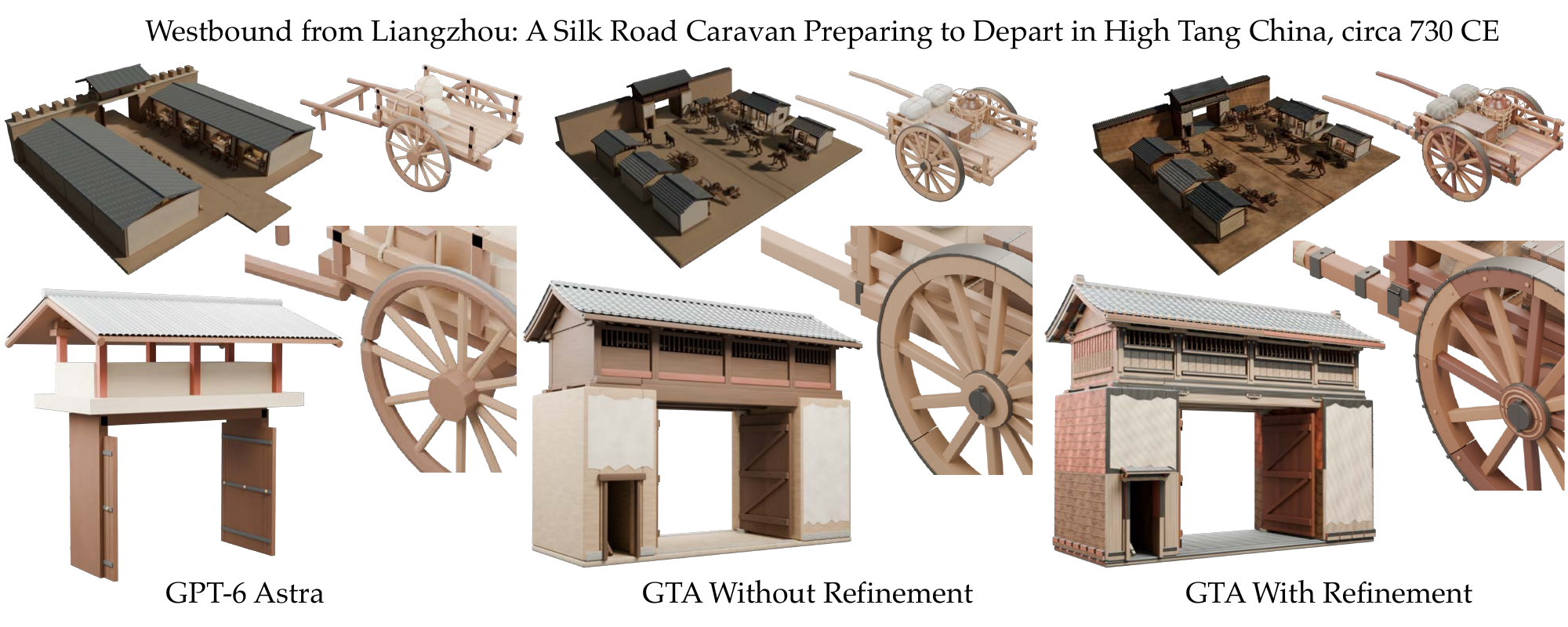}
    \caption{\textbf{GTA refines the details of each object in Stage 3.} }
    \vspace{-0.7cm}
    \label{fig:refinement}
\end{figure}

Fig.~\ref{fig:refinement} compares GTA with and without object-level refinement in Stage 3. Even without refinement, GTA produces more structurally complete objects than Astra. For example, GTA models a detailed gatehouse and side doorway, yielding a more complete structure than Astra’s simple frame. Refinement largely preserves the gate’s structure while making the masonry texture, roof tile edges, and timber framing more distinct. It also highlights the cart wheel’s wood texture and spoke joints. This comparison shows that GTA establishes the main structure of each object before refinement, and that Stage 3 further improves details.

\section{Application}

\begin{figure}[h]
    \centering
    \includegraphics[width=\linewidth]{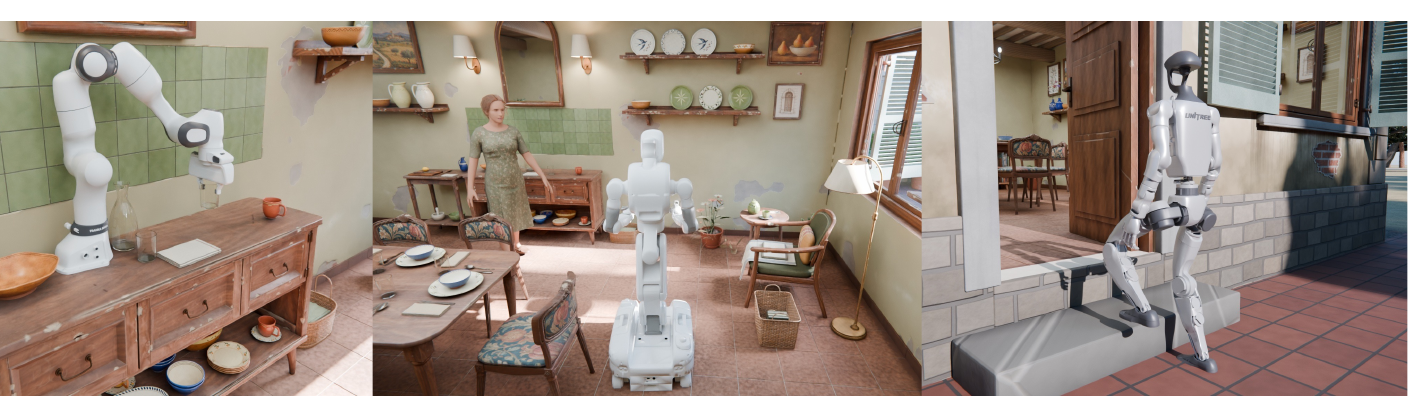}
    \vspace{-0.5cm}
    \caption{\textbf{Task-driven scene generation for robot training.} }
    \vspace{-0.5cm}
    \label{fig:eai}
\end{figure}

\textbf{GTA for game production.} Clear routes and distinct functional areas help players navigate and explore, while detailed modeling of interactable objects makes the scene more convincing up close. Fig.~\ref{fig:addresult_1} shows additional GTA results across diverse regional settings with scene-guided NPCs. \textbf{GTA for robotic training.} GTA can support task-driven robot training by generating functional scenes, as shown in Fig.~\ref{fig:eai}. GTA creates visually diverse environments for diverse robot-training scenarios with context-appropriate human characters. \textbf{Holarchy for other agentic tasks.} While scenes are naturally hierarchical, holarchy can also scale other agentic tasks with a dependency graph by executing ready subtasks in parallel and combining outputs through unified interfaces.

\section{Conclusion}
\label{sec:conclusion}

We introduce GTA, a framework for scalable scene generation in which each holon serves as a scale-agnostic scene unit. This shared representation allows agents to process independent branches concurrently during planning, asset generation, and assembly. GTA generates scenes faster and at higher throughput across diverse types and scales, producing coherent layouts, detailed assets, and fewer object interpenetrations. Scaling studies show that larger scenes benefit more from additional agentic slots. \textbf{Limitations.} GTA tends to generate layouts with distinct spatial groups, making transitions between neighboring regions less gradual, and there is room to improve token usage efficiency.

{
    \small
    \bibliographystyle{ieeenat_fullname}
    \bibliography{main}

@String(CVPR= {IEEE Conf. Comput. Vis. Pattern Recog.})

@String(CVPR  = {CVPR})

@article{paschalidou2021atiss,
  title={Atiss: Autoregressive transformers for indoor scene synthesis},
  author={Paschalidou, Despoina and Kar, Amlan and Shugrina, Maria and Kreis, Karsten and Geiger, Andreas and Fidler, Sanja},
  journal={Advances in neural information processing systems},
  volume={34},
  pages={12013--12026},
  year={2021}
}

@article{jiang2026hog,
  title={HOG-Layout: Hierarchical 3D Scene Generation, Optimization and Editing via Vision-Language Models},
  author={Jiang, Haiyan and Zhang, Deyu and Weng, Dongdong and Song, Weitao and Duh, Henry Been-Lirn},
  journal={arXiv preprint arXiv:2604.10772},
  year={2026}
}

@inproceedings{wang2026worldgen,
  title={Worldgen: From text to traversable and interactive 3d worlds},
  author={Wang, Dilin and Jung, Hyunyoung and Monnier, Tom and Sohn, Kihyuk and Zou, Chuhang and Xiang, Xiaoyu and Yeh, Yu-Ying and Liu, Di and Huang, Zixuan and Nguyen-Phuoc, Thu and others},
  booktitle={Proceedings of the IEEE/CVF Conference on Computer Vision and Pattern Recognition},
  pages={27124--27135},
  year={2026}
}

@inproceedings{yang2024holodeck,
  title={Holodeck: Language guided generation of 3d embodied ai environments},
  author={Yang, Yue and Sun, Fan-Yun and Weihs, Luca and VanderBilt, Eli and Herrasti, Alvaro and Han, Winson and Wu, Jiajun and Haber, Nick and Krishna, Ranjay and Liu, Lingjie and others},
  booktitle={2024 IEEE/CVF Conference on Computer Vision and Pattern Recognition (CVPR)},
  pages={16277--16287},
  year={2024},
  organization={IEEE}
}

@article{xia2026sage,
  title={Sage: Scalable agentic 3d scene generation for embodied ai},
  author={Xia, Hongchi and Li, Xuan and Li, Zhaoshuo and Ma, Qianli and Xu, Jiashu and Liu, Ming-Yu and Cui, Yin and Lin, Tsung-Yi and Ma, Wei-Chiu and Wang, Shenlong and others},
  journal={arXiv preprint arXiv:2602.10116},
  year={2026}
}

@inproceedings{pfaff2026scenesmith,
  title={Scenesmith: Agentic generation of simulation-ready indoor scenes},
  author={Pfaff, Nicholas and Cohn, Thomas and Zakharov, Sergey and Cory, Rick and Tedrake, Russ},
  booktitle={Forty-third International Conference on Machine Learning},
  year={2026}
}

@article{hollis2026sceneagent,
  title={SceneAgent: An Agentic Pipeline from 3D and 4D Captures to Simulation Environments},
  author={Hollis, Luke and Fan, Tianxing and Yang, Heng},
  journal={arXiv preprint arXiv:2610.34745},
  year={2026}
}

@inproceedings{prasad2024adapt,
  title={Adapt: As-needed decomposition and planning with language models},
  author={Prasad, Archiki and Koller, Alexander and Hartmann, Mareike and Clark, Peter and Sabharwal, Ashish and Bansal, Mohit and Khot, Tushar},
  booktitle={Findings of the Association for Computational Linguistics: NAACL 2024},
  pages={4226--4252},
  year={2024}
}

@article{alzu2026roma,
  title={ROMA: Recursive Open Meta-Agent Framework for Long-Horizon Multi-Agent Systems},
  author={Alzu'bi, Salaheddin and Nama, Baran and Kaz, Arda and Eswaran, Anushri and Chen, Weiyuan and Khetan, Sarvesh and Bala, Rishab and Vu, Tu and Oh, Sewoong},
  journal={arXiv preprint arXiv:2602.01848},
  year={2026}
}

@article{wu2023autogen,
  title={Autogen: Enabling next-gen llm applications via multi-agent conversation},
  author={Wu, Qingyun and Bansal, Gagan and Zhang, Jieyu and Wu, Yiran and Li, Beibin and Zhu, Erkang and Jiang, Li and Zhang, Xiaoyun and Zhang, Shaokun and Liu, Jiale and others},
  journal={arXiv preprint arXiv:2308.08155},
  year={2023}
}

@inproceedings{hong2024metagpt,
  title={MetaGPT: Meta programming for a multi-agent collaborative framework},
  author={Hong, Sirui and Zhuge, Mingchen and Chen, Jonathan and Zheng, Xiawu and Cheng, Yuheng and Wang, Jinlin and Zhang, Ceyao and Yau, Steven and Lin, Zijuan and Zhou, Liyang and others},
  booktitle={International Conference on Learning Representations},
  volume={2024},
  pages={23247--23275},
  year={2024}
}

@article{kim2023llm,
  title={An llm compiler for parallel function calling},
  author={Kim, Sehoon and Moon, Suhong and Tabrizi, Ryan and Lee, Nicholas and Mahoney, Michael W and Keutzer, Kurt and Gholami, Amir},
  journal={arXiv preprint arXiv:2312.04511},
  year={2023}
}

@inproceedings{yu2025dyntaskmas,
  title={Dyntaskmas: A dynamic task graph-driven framework for asynchronous and parallel llm-based multi-agent systems},
  author={Yu, Junwei and Ding, Yepeng and Sato, Hiroyuki},
  booktitle={Proceedings of the International Conference on Automated Planning and Scheduling},
  volume={35},
  pages={288--296},
  year={2025}
}

@article{yu2011make,
  title={Make it home: Automatic optimization of furniture arrangement.},
  author={Yu, Lap-Fai and Yeung, Sai Kit and Tang, Chi-Keung and Terzopoulos, Demetri and Chan, Tony F and Osher, Stanley J},
  journal={ACM Trans. Graph.},
  volume={30},
  number={4},
  pages={86},
  year={2011}
}

@inproceedings{tang2024diffuscene,
  title={Diffuscene: Denoising diffusion models for generative indoor scene synthesis},
  author={Tang, Jiapeng and Nie, Yinyu and Markhasin, Lev and Dai, Angela and Thies, Justus and Nie{\ss}ner, Matthias},
  booktitle={2024 IEEE/CVF Conference on Computer Vision and Pattern Recognition (CVPR)},
  pages={20507--20518},
  year={2024},
  organization={IEEE}
}

@article{gao2023scenehgn,
  title={Scenehgn: Hierarchical graph networks for 3d indoor scene generation with fine-grained geometry},
  author={Gao, Lin and Sun, Jia-Mu and Mo, Kaichun and Lai, Yu-Kun and Guibas, Leonidas J and Yang, Jie},
  journal={IEEE Transactions on Pattern Analysis and Machine Intelligence},
  volume={45},
  number={7},
  pages={8902--8919},
  year={2023},
  publisher={IEEE}
}

@inproceedings{lin2024instructscene,
  title={Instructscene: Instruction-driven 3d indoor scene synthesis with semantic graph prior},
  author={Lin, Chenguo and Mu, Yadong},
  booktitle={International Conference on Learning Representations},
  volume={2024},
  pages={25687--25718},
  year={2024}
}

@inproceedings{gao2024graphdreamer,
  title={Graphdreamer: Compositional 3d scene synthesis from scene graphs},
  author={Gao, Gege and Liu, Weiyang and Chen, Anpei and Geiger, Andreas and Sch{\"o}lkopf, Bernhard},
  booktitle={2024 IEEE/CVF Conference on Computer Vision and Pattern Recognition (CVPR)},
  pages={21295--21304},
  year={2024},
  organization={IEEE}
}

@article{chen2023scenedreamer,
  title={Scenedreamer: Unbounded 3d scene generation from 2d image collections},
  author={Chen, Zhaoxi and Wang, Guangcong and Liu, Ziwei},
  journal={IEEE transactions on pattern analysis and machine intelligence},
  volume={45},
  number={12},
  pages={15562--15576},
  year={2023},
  publisher={IEEE}
}

@inproceedings{hu2024scenecraft,
  title={SceneCraft: An LLM Agent for Synthesizing 3D Scenes as Blender Code.},
  author={Hu, Ziniu and Iscen, Ahmet and Jain, Aashi and Kipf, Thomas and Yue, Yisong and Ross, David A and Schmid, Cordelia and Fathi, Alireza},
  booktitle={ICML},
  pages={19252--19282},
  year={2024}
}

@inproceedings{deng2025global,
  title={Global-local tree search in vlms for 3d indoor scene generation},
  author={Deng, Wei and Qi, Mengshi and Ma, Huadong},
  booktitle={2025 IEEE/CVF Conference on Computer Vision and Pattern Recognition (CVPR)},
  pages={8975--8984},
  year={2025},
  organization={IEEE}
}

@article{yang2026sceneweaver,
  title={Sceneweaver: All-in-one 3d scene synthesis with an extensible and self-reflective agent},
  author={Yang, Yandan and Jia, Baoxiong and Zhang, Shujie and Huang, Siyuan},
  journal={Advances in neural information processing systems},
  volume={38},
  pages={140319--140351},
  year={2026}
}

@article{guo2026worldclaw,
  title={Worldclaw: Agentic 3d open-world generation at scale},
  author={Guo, Chunchao and Li, Jinpeng and Li, Yang and Huang, Zilong},
  journal={arXiv preprint arXiv:2608.05248},
  year={2026}
}

@article{li2026hdsl,
  title={HDSL: A Hierarchical Domain-Specific Language for Structured 3D Indoor Scene Generation and Localized Editing with LLM Agents},
  author={Li, Letian and Shen, Chao and Xie, Shuzhao and Gu, Chenghao and He, ZhengXiao and Meng, Yu and Yang, Xin and Jiang, Wenyuan and Wang, Zhi},
  journal={arXiv preprint arXiv:2606.09738},
  year={2026}
}

@article{liu2026scene,
  title={Scene-Designer: Learning Design Intelligence for Agentic 3D Scene Generation},
  author={Liu, Xuanyu and Feng, Hongyan and Pan, Miao and Chen, Sunlai and Liang, Lizhou and Sun, Bing and Zhang, Wenqi and Zhang, Xuhong},
  journal={arXiv preprint arXiv:2610.23121},
  year={2026}
}

@incollection{koestler1969holon,
  author    = {Koestler, Arthur},
  title     = {Beyond Atomism and Holism: The Concept of the Holon},
  booktitle = {Beyond Reductionism: New Perspectives in the Life Sciences},
  editor    = {Koestler, Arthur and Smythies, J. R.},
  publisher = {Hutchinson},
  address   = {London},
  pages     = {192--232},
  year      = {1969}
}

@article{sarandi2024neural,
  title={Neural localizer fields for continuous 3d human pose and shape estimation},
  author={S{\'a}r{\'a}ndi, Istv{\'a}n and Pons-Moll, Gerard},
  journal={Advances in Neural Information Processing Systems},
  volume={37},
  pages={140032--140065},
  year={2024}
}

@inproceedings{pavlakos2019expressive,
  title={Expressive body capture: 3d hands, face, and body from a single image},
  author={Pavlakos, Georgios and Choutas, Vasileios and Ghorbani, Nima and Bolkart, Timo and Osman, Ahmed AA and Tzionas, Dimitrios and Black, Michael J},
  booktitle={Proceedings of the IEEE/CVF conference on computer vision and pattern recognition},
  pages={10975--10985},
  year={2019}
}

@article{rempe2026kimodo,
  title={Kimodo: Scaling controllable human motion generation},
  author={Rempe, Davis and Petrovich, Mathis and Yuan, Ye and Zhang, Haotian and Peng, Xue Bin and Jiang, Yifeng and Wang, Tingwu and Iqbal, Umar and Minor, David and de Ruyter, Michael and others},
  journal={arXiv preprint arXiv:2603.15546},
  year={2026}
}

@article{jiang2026inverse,
  title={Inverse Garment Design with Visual Agents},
  author={Jiang, Nanxiang and Chen, Yunuo and Tao, Zeng and Li, Zizhang and Guo, Qi and Wu, Jiajun and Yang, Yin and Jiang, Chenfanfu},
  journal={arXiv preprint arXiv:2610.15582},
  year={2026}
}

@inproceedings{rosu2025difflocks,
  title={DiffLocks: Generating 3d hair from a single image using diffusion models},
  author={Rosu, Radu Alexandru and Wu, Keyu and Feng, Yao and Zheng, Youyi and Black, Michael J},
  booktitle={2025 IEEE/CVF Conference on Computer Vision and Pattern Recognition (CVPR)},
  pages={10847--10857},
  year={2025},
  organization={IEEE}
}

@inproceedings{xiang2026native,
  title={Native and compact structured latents for 3d generation},
  author={Xiang, Jianfeng and Chen, Xiaoxue and Xu, Sicheng and Wang, Ruicheng and Lv, Zelong and Deng, Yu and Zhu, Hongyuan and Dong, Yue and Zhao, Hao and Yuan, Nicholas Jing and others},
  booktitle={Proceedings of the IEEE/CVF Conference on Computer Vision and Pattern Recognition},
  pages={14419--14429},
  year={2026}
}

@misc{meshy71,
  author       = {{Meshy}},
  title        = {{Meshy API Documentation: Models}},
  howpublished = {\url{https://docs.meshy.ai/en/api/models}},
  note         = {Model identifier: \texttt{meshy-7.1}}
}

@misc{polyhaven,
    author       = {{Poly Haven}},
    title        = {Poly Haven: The Public 3D Asset Library},
    howpublished = {\url{https://polyhaven.com/}},
    note         = {Accessed: 2026-11-16}
}

@misc{polyhaven_sky_fire,
    author       = {{Poly Haven}},
    title        = {The Sky Is On Fire},
    howpublished = {\url{https://polyhaven.com/a/the_sky_is_on_fire}},
    note         = {HDR environment map. Accessed: 2026-11-16}
}

@misc{cgbookcase_birch,
    author       = {{cgbookcase}},
    title        = {Birch Bark 01},
    howpublished = {\url{https://www.cgbookcase.com/textures/birch-bark-01}},
    note         = {PBR texture. Accessed: 2026-11-16}
}

@misc{mapzen_terrain,
    author       = {{Mapzen}},
    title        = {Terrain Tiles},
    howpublished = {\url{https://registry.opendata.aws/terrain-tiles/}},
    note         = {Elevation dataset; Mapzen/Tilezen terrain pipeline.
                    Accessed: 2026-11-16}
}

@book{koestler1967ghost,
  author    = {Koestler, Arthur},
  title     = {The Ghost in the Machine},
  publisher = {Hutchinson},
  address   = {London},
  year      = {1967}
}

@inproceedings{tam2026sceneeval,
  title={SceneEval: Evaluating semantic coherence in text-conditioned 3D indoor scene synthesis},
  author={Tam, Hou In Ivan and Pun, Hou In Derek and Wang, Austin T and Chang, Angel X and Savva, Manolis},
  booktitle={2026 IEEE/CVF Winter Conference on Applications of Computer Vision (WACV)},
  pages={7355--7365},
  year={2026},
  organization={IEEE}
}

@misc{openai_gpt6_astra,
  author       = {{OpenAI}},
  title        = {{GPT-6 Astra}},
  howpublished = {OpenAI API documentation.
                  \url{https://developers.openai.com/api/docs/models/gpt-6-astra}}
}

@misc{openai_gpt_image_15,
  author       = {{OpenAI}},
  title        = {{GPT Image 1.5}},
  howpublished = {OpenAI API documentation.
                  \url{https://developers.openai.com/api/docs/models/gpt-image-1.5}}
}

@misc{openai_gpt61_sol,
  author       = {{OpenAI}},
  title        = {{GPT-6.1 Sol}},
  howpublished = {OpenAI API documentation.
                  \url{https://developers.openai.com/api/docs/models/gpt-6.1-sol}}
}

@misc{anthropic_claude_fable_52,
  author       = {{Anthropic}},
  title        = {System Card for {Claude Fable 5.2} and {Claude Mythos 5.2}},
  year         = {2026},
  howpublished = {\url{https://www.anthropic.com/claude-fable-5-2-mythos-5-2-system-card}}
}

@misc{anthropic_claude_opus_55,
  author       = {{Anthropic}},
  title        = {{Claude Opus 5.5} System Card},
  year         = {2026},
  howpublished = {\url{https://www.anthropic.com/claude-opus-5-5-system-card}}
}
}

\end{document}